\documentclass{article}

\usepackage{PRIMEarxiv}

\usepackage[utf8]{inputenc} 
\usepackage[T1]{fontenc}    
\usepackage{hyperref}       
\usepackage{url}            
\usepackage{booktabs}       
\usepackage{amsfonts}       
\usepackage{nicefrac}       
\usepackage{microtype}      
\usepackage{lipsum}
\usepackage{fancyhdr}       
\usepackage{graphicx}       
\graphicspath{{media/}}     

\usepackage{silence}
\title{VisionPangu: A Compact and Fine-Grained Multimodal Assistant with 1.7B Parameters
}

\author{
  Jiaxin Fan , Wenpo Song\\\\
  National Key Laboratory for Novel Software Technology\\
  Nanjing University \\
}

\begin{document}
\maketitle

\begin{abstract}
Large Multimodal Models (LMMs) have achieved strong performance in
vision-language understanding, yet many existing approaches rely on
large-scale architectures and coarse supervision, which limits their
ability to generate detailed image captions. In this work, we present
VisionPangu, a compact 1.7B-parameter multimodal model designed to
improve detailed image captioning through efficient multimodal
alignment and high-quality supervision. Our model combines an
InternVL-derived vision encoder with the OpenPangu-Embedded language
backbone via a lightweight MLP projector and adopts an
instruction-tuning pipeline inspired by LLaVA. By incorporating dense
human-authored descriptions from the DOCCI dataset, VisionPangu
improves semantic coherence and descriptive richness without relying
on aggressive model scaling. Experimental results demonstrate that
compact multimodal models can achieve competitive performance while
producing more structured and detailed captions. The code and model
weights will be publicly available at \url{https://www.modelscope.cn/models/asdfgh007/visionpangu}.
\end{abstract}


\section{Introduction}
\label{sec:intro}
Humans naturally interact with the world through a tightly integrated combination of vision and language, leveraging complementary modalities to interpret complex visual scenes and communicate nuanced concepts. A long-standing goal in artificial intelligence is to build multimodal assistants capable of aligning visual perception with natural language interaction. While recent advances in language-augmented vision foundation models have enabled strong performance on tasks such as classification, detection, and segmentation, many systems remain constrained by fixed interaction paradigms and struggle to generate detailed visual descriptions in open-ended conversational settings. Despite rapid progress in Large Multimodal Models (LMMs), producing fine-grained, semantically grounded visual narrations remains a challenging problem.

The emergence of Large Language Models (LLMs) has established language as a universal interface for general-purpose AI systems. Through large-scale instruction tuning, models such as ChatGPT and Vicuna demonstrate strong zero-shot generalization across diverse linguistic tasks. Extending this paradigm to multimodal learning has led to early LMMs such as LLaVA and Flamingo, which connect vision encoders with language backbones. However, many existing approaches rely primarily on coarse image-text supervision, which often encourages shallow alignment between visual tokens and language representations. As a result, these models can struggle to produce long-form descriptions that capture fine-grained visual structure and semantic relationships within complex scenes.

In this work, we present a Large Multimodal Model designed to enhance detailed image captioning capabilities within a compact architecture. Our key observation is that dense human-authored visual descriptions provide an implicit form of cross-modal regularization, encouraging the language model to organize visual features into coherent semantic narratives rather than isolated object-level captions. Building on this insight, our model adopts a vision encoder derived from the InternVL3-2B framework, where a pre-trained ViT backbone is separated and further fine-tuned to improve detailed visual representation. This encoder is paired with the OpenPangu-Embedded-1B language model through a lightweight projection module. To leverage recent advances in instruction-following behavior, we follow the LLaVA-NeXT training paradigm and utilize its mixture of multimodal instruction data during both pre-training and supervised fine-tuning (SFT), enabling robust multimodal dialogue grounded in visual content.

A key limitation of many current LMMs lies in their tendency to treat images as collections of independent patches, which can hinder holistic semantic understanding and detailed narration. Rather than focusing solely on model scaling, this work emphasizes improving caption quality through high-quality supervision and encoder adaptation. To this end, we incorporate the DOCCI (Descriptions of Connected Components and Images) dataset, which provides long-form, human-authored, and highly detailed annotations. Fine-tuning on this dataset encourages richer semantic grounding and improves the model’s ability to generate dense, coherent visual narratives aligned with conversational intent.

Our contributions are summarized as follows:
\begin{itemize}
    \item \textbf{Detailed Image Captioning-Oriented LMM:} We develop an end-to-end trained multimodal model tailored for detailed visual narration, combining a ViT encoder adapted from InternVL3-2B with the OpenPangu-Embedded-1B language backbone to achieve strong descriptive capability within a lightweight parameter budget.

    \item \textbf{Instruction Tuning with LLaVA-NeXT:} We adopt the LLaVA-NeXT data mixture and alignment strategy to improve cross-modal instruction following, enabling the model to generalize across diverse visual-language tasks while maintaining efficient training.

    \item \textbf{High-Fidelity Supervision via DOCCI:} By incorporating the DOCCI human-annotated dataset, we push the boundaries of caption quality, enabling the model to generate exhaustive and accurate narrations of complex visual scenes.

    \item \textbf{Technical Performance:} We provide extensive empirical analysis demonstrating that a compact 1.7B-scale multimodal backbone, when paired with a strong vision encoder and high-quality supervision, can achieve competitive performance in multimodal dialogue and dense visual description tasks.
\end{itemize}

\section{Related Work}

\label{sec:related}
\subsection{Vision-Language Pretraining}
Early vision-language research focused on learning joint representations
through large-scale image-text pretraining.
Models such as CLIP~\cite{radford2021clip} and ALIGN~\cite{jia2021align}
demonstrated that contrastive learning on web-scale data can produce
strong cross-modal embeddings.
Subsequent works explored generative pretraining and multimodal fusion
architectures, enabling models to bridge visual perception with natural
language understanding~\cite{li2020oscar, yu2022coca}.
While these approaches achieve strong performance on retrieval and
classification tasks, they typically rely on relatively short captions
and coarse supervision, which can limit their ability to produce
fine-grained visual descriptions.

\subsection{Large Multimodal Models.}
Recent advances in Large Language Models (LLMs) have led to the emergence
of Large Multimodal Models (LMMs), which connect vision encoders with
powerful language backbones.
Flamingo~\cite{alayrac2022flamingo} introduced a gated cross-attention
mechanism to integrate visual tokens into frozen language models,
demonstrating strong few-shot capabilities.
LLaVA~\cite{liu2023visual} and its successors, including
LLaVA-NeXT~\cite{liu2024llavanext}, further showed that instruction tuning
with multimodal conversational data enables flexible visual dialogue.
More recently, InternVL~\cite{chen2024internvl} explored scaling vision
encoders and improving multimodal alignment through large-scale training.
Despite these advances, many LMMs prioritize general reasoning ability,
and relatively fewer works focus explicitly on improving dense visual
narration and detailed image description.

\subsection{Dense Image Description and Captioning.}
Traditional image captioning models such as Show-and-Tell~\cite{vinyals2015show}
and BLIP~\cite{li2022blip} demonstrated that supervised learning on
paired image-text data can produce coherent captions.
More recent datasets emphasize richer and more detailed annotations,
encouraging models to capture fine-grained semantic structure.
In particular, DOCCI~\cite{docci2023dataset} introduces long-form
human-authored descriptions that provide substantially denser supervision
than standard caption datasets.
Our work builds on this line of research by integrating high-fidelity
descriptive data into a modern instruction-tuned LMM, aiming to improve
visual narration without relying on large-scale model scaling.

\section{Methodology}
\label{sec:method}

VisionPangu is designed as a compact yet highly descriptive Large Multimodal Model (LMM). We adopt the end-to-end training paradigm popularized by LLaVA~\cite{liu2023visual}, which connects a pre-trained vision encoder with a large language model (LLM) for general-purpose vision-language interaction. Our approach further adapts this framework to emphasize efficiency and detailed image captioning.

\subsection{Model Architecture}

\begin{figure}[t]
\centering
\includegraphics[width=0.95\linewidth]{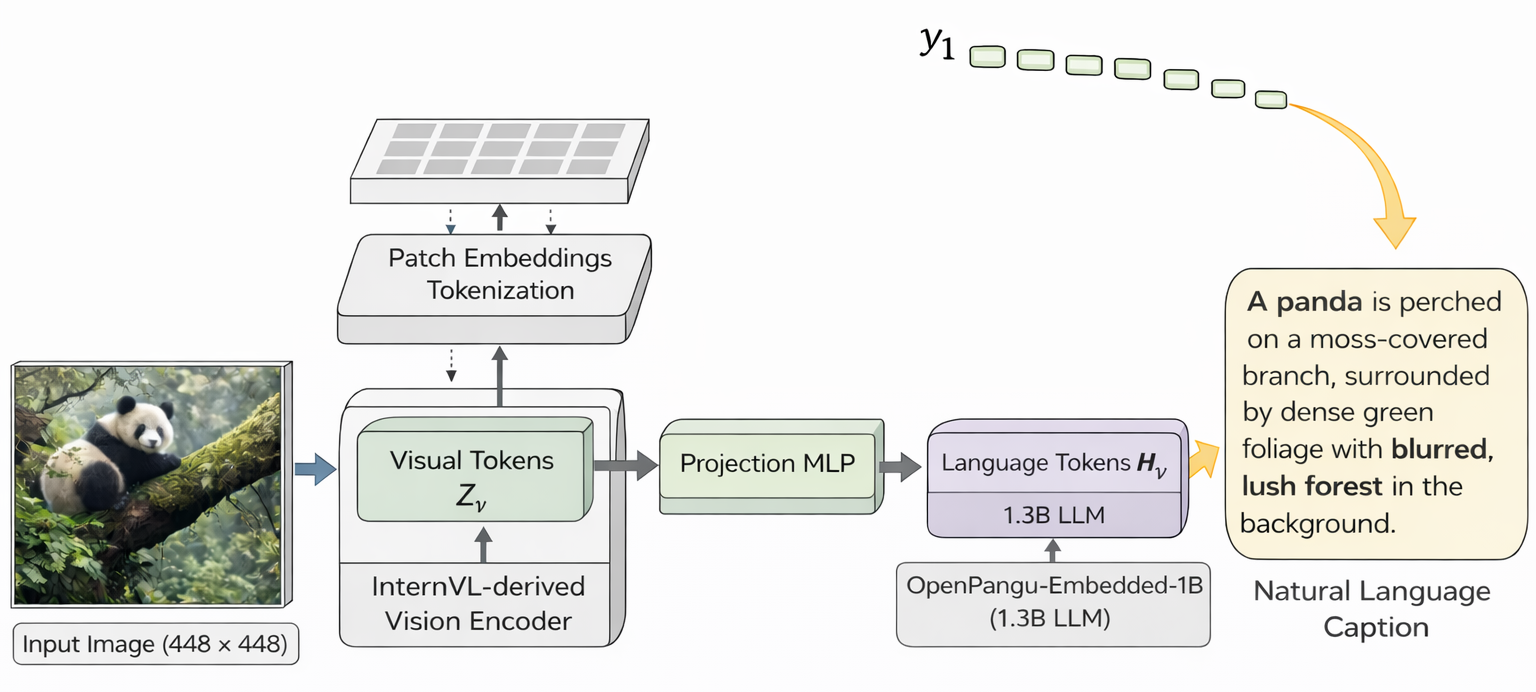}
\caption{
Overview of VisionPangu. The InternVL-derived vision encoder extracts
visual tokens $Z_v$, which are projected into the language embedding
space via a lightweight MLP projector and processed by the
OpenPangu-Embedded-1B language model to generate detailed image captions.
}
\label{fig:architecture}
\end{figure}

The architecture consists of three primary components: a vision encoder, a language decoder, and a projection module that aligns the two modalities.

\textbf{Vision Encoder.}
Instead of directly employing a standalone InternViT model, we derive our visual backbone from the vision encoder of InternVL3-2B~\cite{chen2024internvl}. Specifically, a pre-trained ViT component is separated from the InternVL framework and further fine-tuned to enhance detailed visual representation and dense semantic perception. Compared with the CLIP ViT-L/14 encoder used in the original LLaVA pipeline, this adapted encoder better preserves localized visual structure and high-resolution cues, which are important for fine-grained image description. 
Given an input image $X_v$, the encoder produces visual features
\[
Z_v = g(X_v).
\]

\textbf{Language Model.}
We employ \textbf{OpenPangu-Embedded-1B}~\cite{pangu2024embedded} as the language backbone $f_\phi$. As a lightweight decoder-only transformer, it enables efficient inference while maintaining strong instruction-following capability within a compact parameter scale.

\textbf{Projection Layer.}
To bridge the modality gap between visual features and language embeddings,
we employ a lightweight multi-layer perceptron (MLP) projector rather than a
single linear projection. The MLP consists of stacked fully connected layers
with nonlinear activations, enabling richer cross-modal feature transformation.
Given visual features $Z_v$, the projector maps them into language-aligned
tokens
\[
H_v = \mathrm{MLP}(Z_v),
\]
where $H_v$ shares the same embedding dimensionality as the token space of the
language model.
\subsection{Training Strategy}

Given an input image $X_v$ and a target caption sequence
$\mathbf{y} = (y_1,\ldots,y_T)$, VisionPangu is optimized
under a standard autoregressive multimodal objective.
Specifically, the model maximizes the likelihood of each token
conditioned on previous tokens and projected visual features:

\[
\mathcal{L} = - \sum_{t=1}^{T}
\log p_\phi \big(y_t \mid y_{<t}, H_v \big),
\]

where $H_v=\mathrm{MLP}(g(X_v))$ denotes the visual tokens
produced by the InternVL-derived encoder and projected into
the language embedding space, and $p_\phi$ represents the
language model parameterized by $\phi$.

We follow a two-stage instruction-tuning procedure to ensure effective
cross-modal alignment and strong detailed captioning capability.

\textbf{Stage 1: Pre-training for Feature Alignment.}
In the first stage, the objective is to establish an initial alignment
between visual representations and the language model through a lightweight
MLP projector. We maximize the auto-regressive likelihood of target responses
under the standard instruction-tuning objective. During this stage,
both the vision encoder and the language model remain frozen, and only
the parameters of the MLP projector are updated. We utilize the
LLaVA-NeXT~\cite{liu2024llavanext} pre-training dataset, which reformulates
image-text pairs into a single-turn instruction format to stabilize
early cross-modal alignment.

\textbf{Stage 2: Full-Parameter Instruction Fine-tuning.}
In the second stage, we perform full-parameter fine-tuning of the entire
multimodal model, including the vision encoder, the language model,
and the MLP projector. This stage allows the visual and linguistic
representations to adapt jointly, improving cross-modal interaction
and enabling more detailed caption generation.

To enhance descriptive capability, we adopt a mixed supervision strategy:
\begin{itemize}
\item \textbf{LLaVA-NeXT SFT Mixture:} Maintains general multimodal dialogue
and instruction-following ability~\cite{liu2024llavanext}.
\item \textbf{DOCCI Dataset:} Provides high-fidelity, human-authored
long-form descriptions~\cite{docci2023dataset}, encouraging the model to
capture holistic semantic structure rather than treating images as
isolated patch tokens.
\end{itemize}

\section{Experiments}
\label{sec:exp}

\subsection{Experimental Setup}

\textbf{Training Infrastructure.}
All experiments are conducted on a cluster equipped with eight Huawei
Ascend 910B NPUs (64GB memory per device). Training is performed using
uniform bfloat16 (BF16) precision to ensure stable optimization and
efficient memory utilization on the Ascend platform.

\textbf{Training Strategy.}

We follow the two-stage optimization pipeline described in
Section~\ref{sec:method}. During Stage~1 (feature alignment),
both the vision encoder and the language backbone remain frozen,
and only the parameters of the MLP projector are updated.
This stage initializes the cross-modal mapping between visual
features and language tokens.

During Stage~2 (instruction fine-tuning), we perform full-parameter
fine-tuning of the entire multimodal model, including the vision
encoder, the language model, and the MLP projector. This joint
optimization allows the visual and linguistic representations to
adapt together, improving multimodal reasoning and detailed
caption generation.

\textbf{Data Mixture.}
The training data consists of two components:
(i) the LLaVA-NeXT pre-training and SFT mixtures for general multimodal
instruction following, and (ii) the DOCCI dataset for dense
human-authored visual descriptions. This combination encourages both
robust dialogue ability and fine-grained semantic grounding.

\textbf{Optimization Details.}
Training is conducted using the LLaMA-Factory framework with its
default optimization configuration. The base learning rate is set to
$2\times10^{-5}$, and training is performed in uniform bfloat16 (BF16)
precision on Ascend 910B NPUs. We use a per-device batch size of 8,
resulting in a global batch size of 64 across 8 devices. Images are
resized to a resolution of $448\times448$ before being fed into the
vision encoder. Gradient clipping and cosine learning rate decay are
applied during Stage~2 training.

\subsection{Benchmark Evaluation}

We first evaluate VisionPangu on several standard multimodal benchmarks,
including MME, MMMU, and POPE, to assess its general multimodal reasoning
and instruction-following capabilities. Although VisionPangu is primarily
designed for detailed image captioning rather than large-scale multimodal
reasoning, the results show that a compact 1.7B-parameter architecture can
still achieve competitive performance across commonly used benchmarks.

Table~\ref{tab:benchmark_eval} presents the comparison with several
representative lightweight multimodal models. Despite its relatively
small parameter scale, VisionPangu achieves comparable performance to
existing models in the same parameter range. These results suggest that
efficient architecture design and high-quality supervision can partially
compensate for the lack of aggressive model scaling.

\begin{table}[t]
\centering
\caption{Benchmark comparison with representative multimodal models.
Higher is better for all metrics.}
\begin{tabular}{lcccc}
\toprule
Model & MMMU & MMbench & POPE & MME \\
\midrule
VisionPangu-1.7B (Ours) & 36.5 & 62.5 & 82.8 & 283.21 / 1279.39 \\
InternVL2-2B              & 36.3 & 73.2 & 87.0 & 1876.8 (sum) \\
Qwen2-VL-2B               & 41.1 & 74.9 & 87.6 & 413.9 / 1474.1 \\
MiniCPM-V 2.0 (2B)        & 38.2 & 69.6 & 86.3 & 396.8 / 1411.4 \\
Bunny-v1.1-Llama3-8B      & 43.3 & 77.2 & 87.2 & 367.5 / 1644.1 \\
LLaVA-v1.6-Vicuna-7B      & 35.8 & 67.4 & 86.5 & 332 / 1519 \\
\bottomrule
\end{tabular}
\label{tab:benchmark_eval}
\end{table}

\subsection{Captioning Evaluation}

Beyond general multimodal benchmarks, we further evaluate VisionPangu
on detailed image captioning tasks to measure its ability to produce
fine-grained and long-form visual descriptions. Unlike traditional
short-caption benchmarks, detailed captioning emphasizes semantic
richness, descriptive completeness, and coherent narrative structure.

To evaluate detailed captioning quality, we construct an evaluation
subset from the COCO 2017 validation set. Specifically, we randomly
sample 600 images and generate captions using each model for comparison.
This subset allows us to measure descriptive richness and semantic
consistency while maintaining a manageable evaluation scale.

We report results using widely adopted captioning metrics, including
BLEU, METEOR, and ROUGE-L, which provide quantitative measurements of
caption quality from complementary perspectives.

\subsection{Captioning Benchmark Comparison}

To further evaluate the descriptive capability of VisionPangu, we
compare our model with several representative lightweight multimodal
models on detailed captioning benchmarks.

Table~\ref{tab:caption_eval} summarizes the quantitative results.
VisionPangu achieves the best performance across all reported metrics,
including BLEU, METEOR, and ROUGE-L. In particular, the model shows a
substantial improvement in BLEU and ROUGE-L scores compared with
other models of similar scale, indicating stronger ability to generate
structured and semantically rich long-form descriptions.

These results highlight the effectiveness of combining dense caption
supervision with efficient multimodal alignment, enabling compact
multimodal models to achieve strong performance on detailed captioning
tasks.

\begin{table}[t]
\centering
\caption{Detailed image captioning evaluation. }
\begin{tabular}{lccccc}
\toprule
Model & BLEU & METEOR & ROUGE-L  \\
\midrule
VisionPangu-1.7B (Ours) &\textbf{0.2859}& \textbf{0.4708} &\textbf{0.3759}  \\
InternVL2-2B             & 0.0954 & 0.3862 &  0.2473   \\
Qwen2-VL-2B              &0.1230 &0.4352  & 0.2531  \\
MiniCPM-V 2.0 (2B)        & 0.1198 &0.4484  & 0.2520 &   \\
LLaVA-v1.6-Vicuna-7B     & 0.2431& 0.4329&  0.3092    \\

\bottomrule
\end{tabular}
\label{tab:caption_eval}
\end{table}




\section{Conclusion}
\label{sec:conclusion}

In this work, we introduced VisionPangu, a compact 1.7B-parameter
multimodal model designed to enhance detailed image captioning through
efficient multimodal alignment and high-fidelity supervision.
By combining an InternVL-derived vision encoder, the
OpenPangu-Embedded language backbone, and a lightweight MLP projector,
our architecture achieves strong descriptive capability while
maintaining a relatively small computational footprint.

Through instruction tuning with the LLaVA-NeXT data mixture and
additional supervision from the DOCCI dataset, VisionPangu improves
long-form caption generation and semantic coherence without relying on
aggressive model scaling. Experimental results across multimodal
benchmarks and captioning evaluations demonstrate that compact models
can remain competitive when paired with high-quality training signals
and efficient design choices.

Future work will explore higher-resolution visual representations,
more robust evaluation protocols for detailed captioning, and
extensions toward video understanding and multi-image reasoning,
further advancing practical and efficient multimodal assistants.

\bigskip

\bibliographystyle{unsrt}  
\bibliography{references}

\end{document}